\documentclass[Afour,sageh,times]{sagej}

\usepackage{moreverb,url}
\usepackage[colorlinks,bookmarksopen,bookmarksnumbered,citecolor=red,urlcolor=red]{hyperref}

\newcommand\BibTeX{{\rmfamily B\kern-.05em \textsc{i\kern-.025em b}\kern-.08em
T\kern-.1667em\lower.7ex\hbox{E}\kern-.125emX}}

\def\volumeyear{2019}

\usepackage{tabularx}
\usepackage{booktabs}
\usepackage{multirow,array}
\usepackage{gensymb}
\usepackage{adjustbox}

\usepackage{amsmath} % assumes amsmath package installed
\usepackage{amssymb}  % assumes amsmath package installed

\usepackage{forest}
\usepackage{subcaption}

\RequirePackage[utf8]{inputenc}

\begin{document}

\runninghead{Ferrera et al.}

\title{AQUALOC: An Underwater Dataset for Visual-Inertial-Pressure Localization.}

\author{Maxime Ferrera\affilnum{1,2}, Vincent Creuze\affilnum{2}, Julien Moras\affilnum{1} and Pauline Trouv\'e-Peloux\affilnum{1}}

\affiliation{
\affilnum{1} DTIS, ONERA, Universit\'e Paris Saclay F-91123 Palaiseau, France\\
\affilnum{ 2} LIRMM, Univ. Montpellier, CNRS, Montpellier, France
}

\corrauth{Maxime Ferrera}

\email{maxime.ferrera@gmail.com}

\begin{abstract}
\textit{We present a new dataset, dedicated to the development of simultaneous localization and mapping methods for underwater vehicles navigating close to the seabed.  The data sequences composing this dataset are recorded in three different environments: a harbor at a depth of a few meters, a first archaeological site at a depth of 270 meters and a second site at a depth of 380 meters.  The data acquisition is performed using Remotely Operated Vehicles equipped with a monocular monochromatic camera, a low-cost inertial measurement unit, a pressure sensor and a computing unit, all embedded in a single enclosure.  The sensors' measurements are recorded synchronously on the computing unit and seventeen sequences have been created from all the acquired data.  These sequences are made available in the form of ROS bags and as raw data.  For each sequence, a trajectory has also been computed  offline using a Structure-from-Motion library in order to allow the comparison with real-time localization methods.  With the release of this dataset, we wish to provide data difficult to acquire and to encourage the development of vision-based localization methods dedicated to the underwater environment.  The dataset can be downloaded from: \url{http://www.lirmm.fr/aqualoc/}}

\end{abstract}


\maketitle

\section{Introduction}

Accurate localization is critical for mobile robotics.  In open outdoor areas, it can be obtained from Global Positioning System (GPS).  However, in GPS-denied environments, such as indoor or beneath the sea surface, robots' position must be estimated from other sensors.

In underwater robotics, the localization problem is often solved by coupling high-grade Inertial Measurement Units (IMU) with compass, Doppler Velocity Logs (DVL) and pressure sensors (\cite{Paull2014}).  Such solutions, classified as dead-reckoning (DR) localization, are highly dependent of the sensors quality and suffer from unbounded drift.  While these methods can be employed quite safely for vehicles navigating in the middle of the water column (\textit{i.e.} in obstacle free areas), they are not accurate enough for navigation in cluttered areas.  In such places, Simultaneous Localization And Mapping (SLAM) methods are preferred.  SLAM requires exteroceptive sensors, such as Lidar, sonar or camera, to measure the 3D structure of the environment.  From these data, the localization is estimated while a 3D map is progressively built.

Visual SLAM (VSLAM) and Visual-Inertial Odometry (VIO) have been a hot research topic during the past decades (\cite{PastPresentFutureSLAM}).  VSLAM consists in estimating localization from visual data, possibly enhanced by complementary sensors, through the mapping of the observed scenes.  In ground and aerial robotics, the availability of many public datasets, such as KITTI (\cite{kitti}), Malaga (\cite{Malaga2014}) or EuRoC (\cite{Euroc}), to cite a few, has greatly impacted the development of VSLAM methods.  Recent algorithms, relying on monocular cameras (\cite{ORB-SLAM, SVO-2, DSO}) or on visual-inertial sensors (\cite{okvis,Mur_Artal_2017,vinsmono}), have shown impressive results, with centimetric localization accuracy. 
In underwater robotics, many operations occur near the seabed (biology, Oil\&Gas Industry, mine warfare, archaeology...), making visual information available.  Nonetheless, in such conditions, the acquired images suffer from degradation like turbidity, backscattering and illumination issues, due to the medium properties.  These poor imaging conditions must be accounted for in the development of underwater VSLAM or VIO systems, thus preventing use of the previously cited algorithms (\cite{Li_2016,uwstereomap,KneipUWodo}).  Some previous works have investigated the use of monocular camera for underwater localization (\cite{TrajBasedVisualSLAM,ferrera2019}), sometimes coupled to low-cost IMU and pressure sensor (\cite{ShkurtiEKF-SLAM, CreuzeMonoVO}), sonars (\cite{svin}) or even as a mean of detecting loop-closures in DR systems (\cite{VisualSaliencySLAM}).  However, the limited amount of public datasets dedicated to this localization challenge prevent a fair comparison of these methods on common data.  Moreover, the fact that these data are difficult to acquire, because of the required equipment and logistic, limits the development of new methods.    
\cite{marinecfr} proposed a dataset containing the measurements of navigational sensors, stereo cameras and a multibeam sonar.  \cite{cavesonar} released another dataset dedicated to localization and mapping in an underwater cave from sonar measurements.  Images acquired by a monocular camera are also given for the detection of cones precisely placed in order to have a mean of estimating drift.  However, in both datasets, the acquisition rate of the cameras is too low ($<$10 Hz) for most of VSLAM and VIO methods.  \cite{UWsimData} created a synthetic dataset simulating the navigation of a vehicle in an underwater environment and containing monocular cameras measurements at a frame-rate of 10 Hz.  Many public datasets have also been made available by the Oceanography community through national websites (\url{https://www.data.gov/}, \url{http://www.marine-geo.org}).  However, these datasets have not been gathered with the aim of providing data suitable for VSLAM or VIO and often lack essential information such as the calibration of their sensors' setup.
% Yet, to the best of our knowledge, there is no underwater dataset providing real data dedicated to the development of VSLAM algorithms.

In this paper, we present AQUALOC, a new dataset aiming at the development of VSLAM and VIO methods dedicated to the underwater environment.  The dataset sequences have been recorded using acquisition systems composed of a monochromatic camera, a Micro Electro-Mechanical System (MEMS) based IMU, a presssure sensor and a computing unit for synchronous recordings.  These acquisition systems have been embedded on ROVs equipped with lighting systems and navigating close to the seabed.  The recorded video sequences exhibit the typical visual degradation induced by the underwater environment such as turbidity, backscattering, shadows and strong illumination shifts caused by the artificial lighting systems.  Three different sites have been explored to create the dataset: a harbor and two archaeological sites.  The recording of the sequences occurred at different depths, going from a few meters, for the harbor, to several hundred meters, for the archaeological sites.  The provided video sequences are hence highly diversified in terms of scenes (low-textured areas, very texture repetitive areas...) and of scenarios (exploration, photogrammetric surveys, manipulations...).  As the acquisition of ground truth is very difficult in natural underwater environments, we have used the state-of-the-art Structure-from-Motion (SfM) library Colmap (\cite{Colmap}) to compute comparative baseline trajectories for each sequence.  Colmap processes offline the sequences and performs a 3D reconstruction to estimate the positions of the camera.  This 3D reconstruction is done by matching exhaustively all the images composing a sequence, which allows the detection of many loop-closures and, hence, the computation of accurate trajectories, assessed by low average reprojection errors.  Along with the computed trajectories, we also provide the list of matched images for each sequence which could be used to evaluate relocalization or loop-closure detection methods.  We further include statistics on the 3D reconstruction to assess their accuracy.

With the release of this dataset, we provide to the community the opportunity to work on data difficult to acquire.  Indeed, the logistic (ship availability) and the required equipment (deep-sea compliant underwater vehicles and sensors), as well as regulations (official authorizations), can be a barrier preventing possible works on this topic.  We are convinced that the availability of this dataset will increase the development of algorithms dedicated to the underwater environment.  Both raw and ROS bag formatted field data are provided along with the full calibration of the sensors (camera and IMU).  Moreover, the provided comparative baseline makes this dataset suitable for benchmarking VSLAM and VIO algorithms.

The rest of this paper is organized as follows.  First, we present the design of the acquisition systems used and the calibration procedures employed.  Then, an overview of the dataset is given and the acquisition conditions on each site are detailed, highlighting the associated challenges for visual localization.  Next, the processing of the data sequences to create a baseline is described.  Finally, we detail how the dataset is organized and in which way the data are formatted.

\section{The Acquisition Systems}
\label{sec:acq_sys}

\begin{figure}
    \centering
        \begin{subfigure}[b]{0.4625\linewidth}
            \centering \includegraphics[width=\linewidth]{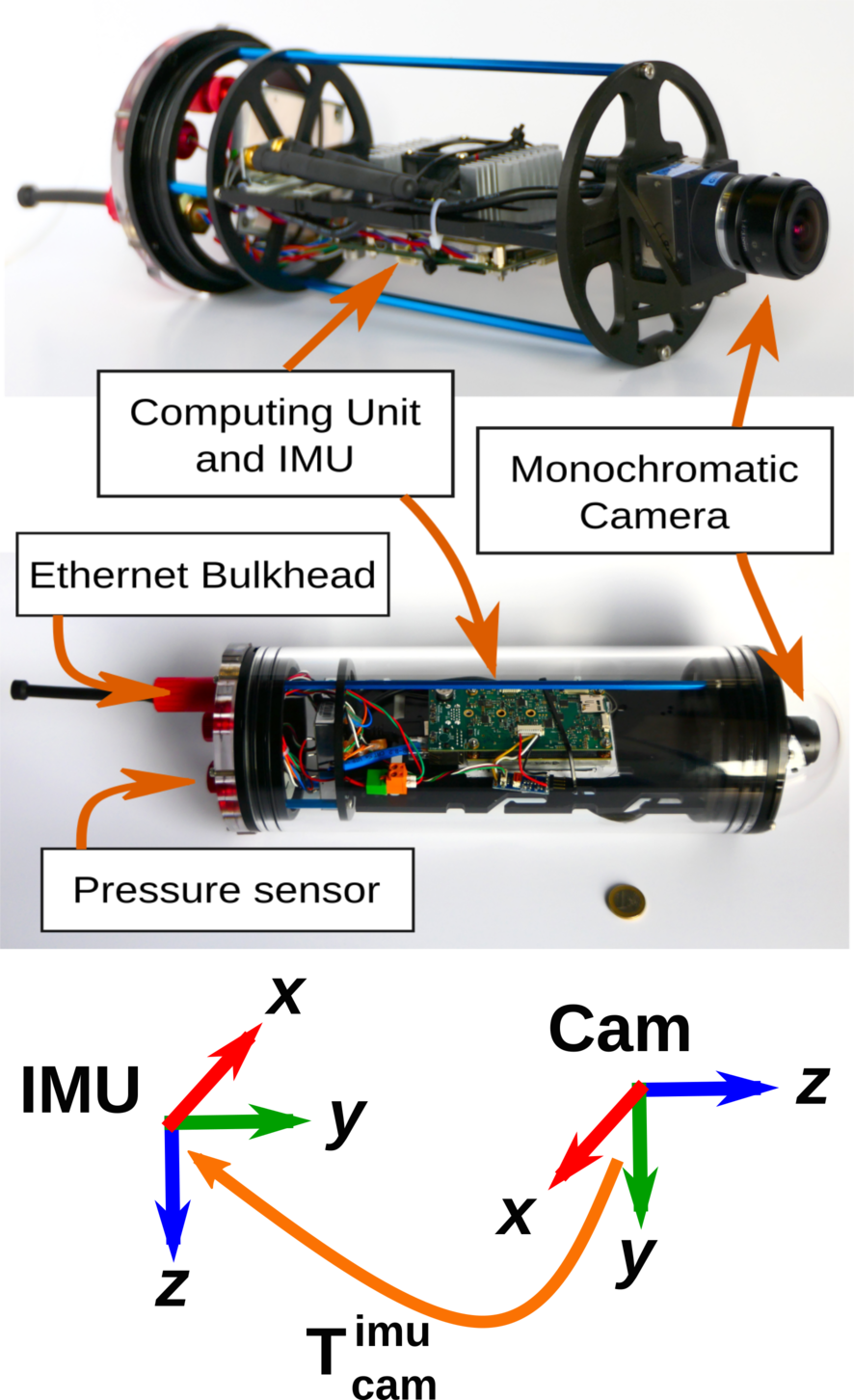}
            \captionsetup{justification=centering}
            \caption{System A}\label{fig:caissonA}
        \end{subfigure}
        ~
        \begin{subfigure}[b]{0.4375\linewidth}
            \centering \includegraphics[width=\linewidth]{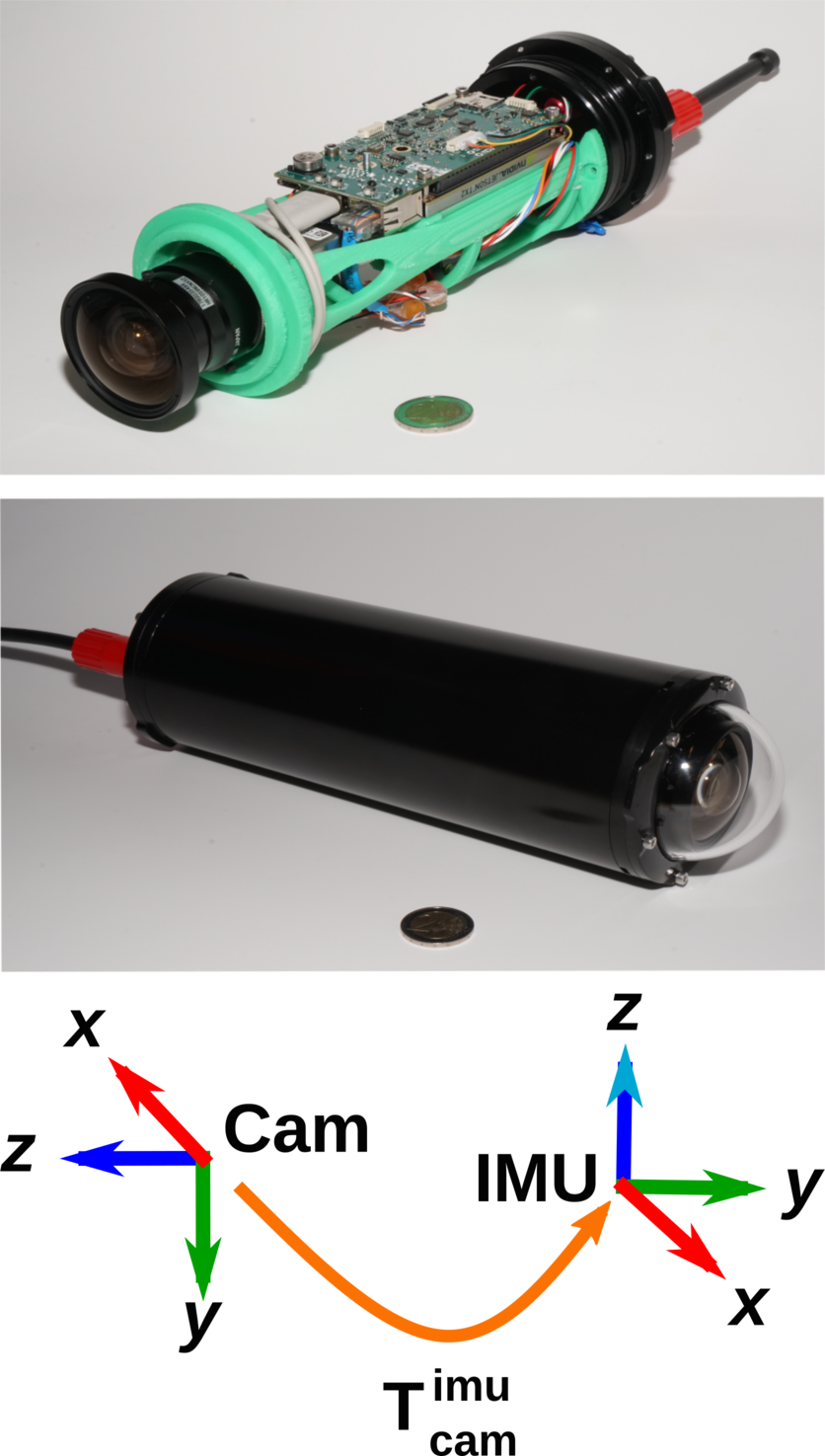} 
            \captionsetup{justification=centering}
            \caption{System B}\label{fig:caissonB}
        \end{subfigure}
    \caption{The acquisition systems equipped with a monocular monochromatic camera, a pressure sensor, an IMU and a computer along with the sensors' reference frames.}
    \label{fig:caissons}
\end{figure}

In order to acquire the sequences of the dataset, we have designed two similar underwater systems.  These acquisition systems have been designed to allow the localization of underwater vehicles from a minimal set of sensors in order to be as cheap and as versatile as possible.  Both systems are equipped with a monochromatic camera, a pressure sensor, a low-cost MEMS-IMU and an embedded computer.  The camera is placed behind an acrylic dome to minimize the distortion effects induced by the difference between water and air refractive indices.  The image acquisition rate is 20 Hz.  The IMU delivers measurements from a 3-axes accelerometer, 3-axes gyroscope and 3-axes magnetometer at 200 Hz.  The embedded computer is a Jetson TX2 running Ubuntu 16.04 and is used to record synchronously the sensors' measurements thanks to the ROS middleware.  The Jetson TX2 is equipped with a carrier board embedding the mentioned MEMS-IMU and a 1 To NVME SSD to directly store the sensors measurements, thus avoiding any bandwidth or package loss issue.  An advantage of the self-contained systems that we have developed, is that they are independent of any robotic architecture and can thus be embedded on any kind of Remotely Operated Vehicle (ROV) or Autonomous Underwater Vehicle (AUV).  The interface can either be Ethernet or a serial link, depending on the host vehicle's features.  

To record data at different depths, we have designed two systems that we will refer to as ``System A'' and ``System B''.  These systems have the same overall architecture, but they differ on the camera model, the pressure sensor type and the diameter and material of the enclosure.  System A (Fig.~\ref{fig:caissonA}) is designed for shallow waters and was used to acquire the sequences in the harbor.  Its camera has been equipped with a wide-angle lens, which can be modelled by the fisheye distortion model.  The pressure sensor is rated for 30 bars and delivers depth measurements at a maximum rate of 10~Hz.  System A is protected by an acrylic enclosure, rated for a depth of 100 meters.  System B (Fig.~\ref{fig:caissonB}) was used to record the sequences on the archaeological sites at larger depths.  Its camera has a slightly lower field of view and the lens can be modelled by the radial-tangential distortion model.  It embeds a pressure sensor rated for 100 bars delivering depth measurements at 60 Hz.  Its enclosure is made of aluminum and is 400 meters depth rated.  The technical details about both systems and their embedded sensors are given in table~\ref{tab:sensors}.

Each camera-IMU setup has been cautiously calibrated to provide the intrinsic and extrinsic parameters required to use it for localization purpose.  We have used the toolbox Kalibr (\cite{Kalibr_cam, Kalibr_imucam}) along with an apriltag target to compute all the calibration parameters.

The cameras calibration step allows to obtain an estimate of the focal lengths, principal points and distortion coefficients.  These parameters can then be used to undistort the captured images and to model the image formation pipeline, with the following notation:

\begin{gather}
\begin{bmatrix} u \\ v \end{bmatrix} = \Pi_{\mathbf{K}}  \left ( \mathbf{R}^{\text{cam}}_{\text{w}} \mathbf{X}_{\text{w}} +  \mathbf{t}^{\text{cam}}_{\text{w}} \right )
\end{gather}

\begin{gather}
\begin{bmatrix} u \\ v \end{bmatrix} = \begin{bmatrix} f_{x}.\frac{x_{\text{cam}}}{z_{\text{cam}}} + c_x  \\ f_{y}.\frac{y_{\text{cam}}}{z_{\text{cam}}} + c_y \end{bmatrix} = \Pi_{\mathbf{K}} \left ( \mathbf{X}_{\text{cam}} \right )
\end{gather}

\begin{gather*}
\textup{with } \mathbf{K} = \begin{bmatrix} f_x & 0 & c_x \\ 0 & f_y & c_y \\ 0 & 0 & 1 \end{bmatrix} \textup{ and } 
\mathbf{X}_{\text{cam}} = \begin{bmatrix} x_{\text{cam}} \\ y_{\text{cam}} \\ z_{\text{cam}} \end{bmatrix}
\end{gather*}

\noindent where $\Pi_{\mathbf{K}} (\cdot)$ denotes the projection: $\mathbb{R}^{3} \mapsto \mathbb{R}^{2}$, $\mathbf{K}$ is the calibration matrix, $\mathbf{X}_{\text{w}} \in \mathbb{R}^{3}$ is the position of a 3D landmark in the world frame, $\mathbf{R}^{\text{cam}}_{\text{w}} \in SO(3)$ and $\mathbf{t}^{\text{cam}}_{\text{w}} \in \mathbb{R}^{3}$ denote the rotational and translational components of the transformation from the world frame to the camera frame,  $\mathbf{X}_{\text{cam}} \in \mathbb{R}^{3}$ is the position of a 3D landmark in the camera frame, $f_x$ and $f_y$  denotes the focal lengths and $(c_x,c_y)$ is the principal point of the camera.

\begin{figure}
\centering
\includegraphics[width=\linewidth]{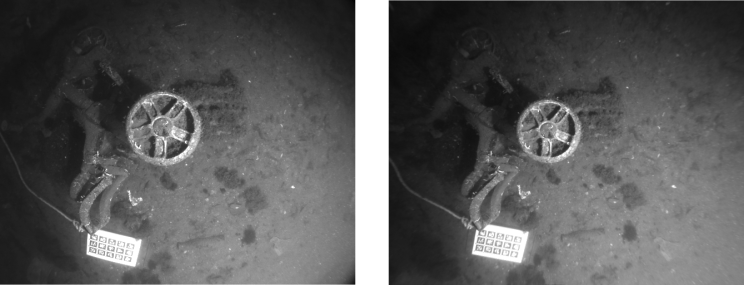}
\caption{Distortion effects removal from Kalibr calibration on one of the harbor sequences. Left: raw image.  Right: undistorted image.}
\label{fig:kalibr}
% \vspace{-7.5mm}
\end{figure}

As these parameters are medium dependant, the calibration has been performed in water to account for the additional distortion effects at the dome's level.  The results of the calibration of the fisheye camera can be seen in figure \ref{fig:kalibr}.

The camera-IMU setup calibration allows to estimate the extrinsic parameters of the setup, that is the relative position of the camera with respect to the IMU, and the time delay between camera's and IMU's measurements.  This relative position is represented by a rotation matrix $\mathbf{R}^{\text{imu}}_{\text{cam}} $ and a translation vector $\mathbf{t}^{\text{imu}}_{\text{cam}}$.  Camera and IMU's poses relate to each other through:

\begin{gather}
\mathbf{T}^{\text{w}}_{\text{cam}} = \mathbf{T}^{\text{w}}_{\text{imu}} \mathbf{T}^{\text{imu}}_{\text{cam}}
\end{gather}

\begin{gather*}
\textup{with } \mathbf{T}^{\text{imu}}_{\text{cam}} \doteq
\begin{bmatrix}
\mathbf{R}^{\text{imu}}_{\text{cam}} & \mathbf{t}^{\text{imu}}_{\text{cam}} \\ 
0_{1\times3} & 1
\end{bmatrix} \in \mathbb{R}^{4\times4}
\\ \\
\textup{and } \left ( \mathbf{T}^{\text{w}}_{\text{cam}} \right )^{-1} = \mathbf{T}^{\text{cam}}_{\text{w}} \doteq
\begin{bmatrix}
\mathbf{R}^{\text{cam}}_{\text{w}} & \mathbf{t}^{\text{cam}}_{\text{w}} \\ 
0_{1\times3} & 1
\end{bmatrix} \in \mathbb{R}^{4\times4}
\end{gather*}

\noindent where $\mathbf{R}^{\text{imu}}_{\text{cam}}  \in SO(3)$, $\mathbf{t}^{\text{imu}}_{\text{cam}} \in \mathbb{R}^{3}$, $\mathbf{T}^{\text{w}}_{\text{cam}} \in SE(3)$, $\mathbf{T}^{\text{cam}}_{\text{w}} \in SE(3)$, $\mathbf{T}^{\text{imu}}_{\text{cam}} \in SE(3)$ and $\mathbf{T}^{\text{w}}_{\text{imu}} \in SE(3)$.  $\mathbf{T}^{\text{w}}_{\text{cam}}$ and $\mathbf{T}^{\text{w}}_{\text{imu}}$ respectively represent the poses of the camera and of the body, with respect to the world frame.  $\mathbf{T}^{\text{cam}}_{\text{w}}$ is the inverse transformation of $\mathbf{T}^{\text{w}}_{\text{cam}}$ and $\mathbf{T}^{\text{imu}}_{\text{cam}}$ is the transformation from the camera frame to the IMU frame. 

Before estimating these extrinsic parameters, the IMU noise model parameters have been derived from an Allan standard deviation plot, obtained by recording the gyroscope and accelerometer measurements for several hours, while keeping the IMU still.  These noise parameters are then fed into the calibration algorithms to model the IMU measurements.  As these parameters (IMU noises, camera-IMU relative transformation and measurements time delay) are independent of the medium (air or water), they have been estimated in air.  Doing this calibration step in air allowed to perform easily the fast motions required to correlate the IMU measurements to the camera's ones.

All the calibration results are included in the dataset, that is the cameras' models (including the intrinsic parameters and the distortion coefficients), the IMUs' noise parameters, the relative transformation from the camera to the IMU and the time delay between the cameras' and the IMUs' measurements.

\begin{table}
\centering
\begin{adjustbox}{width=\linewidth}
\begin{tabular}{@{}cll@{}}
\toprule

\multirow{17}{*}{\shortstack{System A \\ (Harbor \\ sequences)}} 		  
          & \textbf{Camera sensor}             & \textbf{UEye - UI-1240SE}          \\
          & Resolution                         & 640$\times$512 px      \\
          & Sensor                             & Monochromatic         \\
          & Frames per second                   & 20 fps                \\
          & \textbf{Lens}                      & \textbf{Kowa LM4NCL C-Mount}	   \\
          & Focal length                       & 3.5mm          \\
          & \textbf{Pressure Sensor}           & \textbf{MS5837 - 30BA}             \\
          & Depth range                        & 0 - 290m       \\
          & Resolution                         & 0.2 mbar                  \\
          & Output frequency                   & 5-10 Hz                \\
          & \textbf{Inertial Measurement Unit} & \textbf{MEMS - MPU-9250}           \\
          & Gyroscope frequency                & 200 Hz      \\
          & Accelerometer frequency            & 200 Hz                              \\
          & Magnetometer frequency             & 200 Hz                              \\
          & \textbf{Embedded Computer}         & \textbf{Nvidia - Tegra Jetson TX2}  \\
          & Carrier board                      & Auvidea J120 - IMU                  \\
          & Storage                            & NVME SSD 1 To                          \\
          & \textbf{Housing}                   & \textbf{4" Blue Robotics Enclosure}  \\
   		  & Enclosure						   & 33.4 x 11.4 cm    \\
   		  & Enclosure Material				   & Acrylic         \\
  		  & Dome            				   & 4" Blue Robotics Dome End Cap    \\ 
  \midrule

\multirow{17}{*}{\shortstack{System B \\ (Archaeo. \\ sequences)}} 		  
          & \textbf{Camera sensor}             & \textbf{UEye - UI-3260CP }   \\
          & Resolution                         & 968$\times$608 px   \\
          & Sensor                             & Monochromatic       \\
          & Frames per second                  & 20 fps           \\
          & \textbf{Lens}                      & \textbf{Kowa LM6NCH C-Mount}	\\
          & Focal length                       & 6mm                    \\
          & \textbf{Pressure Sensor}           & \textbf{Keller 7LD - 100BA}     \\
          & Depth range                        & 0 - 990m              \\
          & Resolution                         & 3 mbar                 \\
          & Output frequency                   & 60 Hz                  \\
          
          & \textbf{Inertial Measurement Unit} &   \textbf{\textit{Same as System A}} \\
          
          & \textbf{Embedded Computer}   & \textbf{\textit{Same as System A}}          \\
          
          & \textbf{Housing}            & \textbf{3" Blue Robotics Enclosure}  \\
   		  & Enclosure				    & 25.8 x 8.9 cm                  \\
   		  & Enclosure Material	        & Aluminium    \\
  		  & Dome            			& 3" Blue Robotics Dome End Cap 
\\ \cmidrule(l){1-3} 
\end{tabular}
\end{adjustbox}
\captionsetup{justification=centering}
\caption{Technical details about the acquisition systems.}
\label{tab:sensors}
\end{table}

\section{Dataset Overview}

As explained in section~\ref{sec:acq_sys}, System A was used to record the shallow harbor sequences, while System B was used on the two deep archaeological sites.  We propose a total of 17 sequences: 7 recorded in the harbor, 4 on the first archaeological site and 6 on the second site.  As each of these environments is in some ways different from the others, we describe the sequences recorded in each environment separately.  Table \ref{tab:sequences} summarizes the specificities of each data sequence.  Note that, for each sequence, the starting and ending points are approximately the same.  In most of the sequences, there are closed loops along the performed trajectories.  Some sequences also slightly overlap, which can be useful for the development of relocalization features.

\subsection{Harbor sequences}
\label{sec:harbor}

\begin{figure}
\centering
\includegraphics[width=0.8\linewidth]{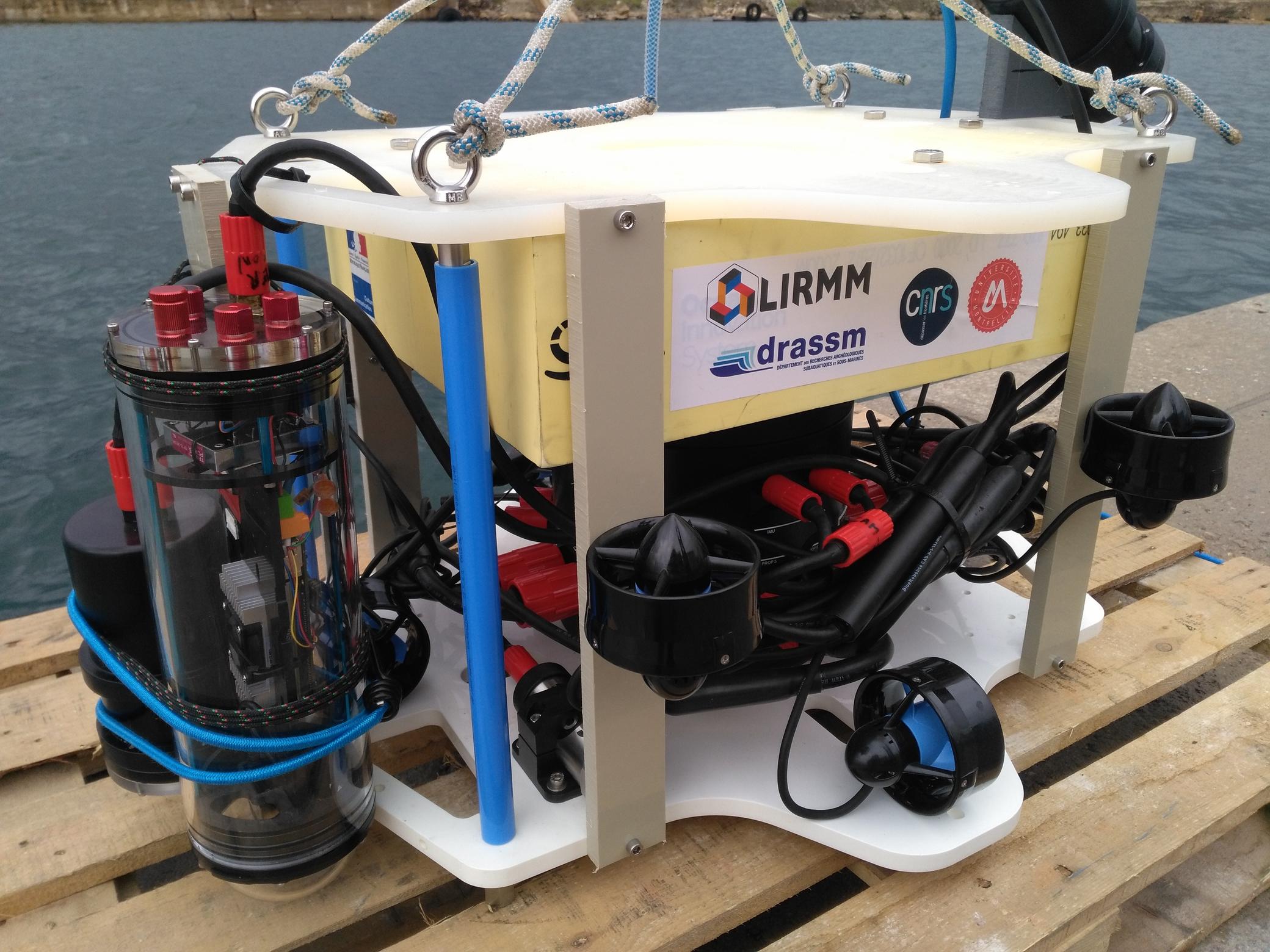}
\caption{The Remotely Operated Vehicle \textit{Dumbo} and the acquisition system A, used to record the harbor sequences.}
\label{fig:rov_and_caisson}
\end{figure}

The harbor sequences were recorded in April 2018.  System A was embedded on the lightweight ROV \textit{Dumbo} (\textit{DRASSM-LIRMM}) with the camera facing downward, as shown in figure \ref{fig:rov_and_caisson}.  The ROV was navigating at a depth of 3 to 4 meters over an area of around $100$ m$^{2}$.  Although the sun illuminates this shallow environment, a lighting system was used in order to increase the signal-to-noise ratio of the images acquired by the camera.  The explored area was mostly planar but the presence of several big objects made it a real 3D environment, with significant relief.

For each sequence, loops are performed and an apriltag calibration target is used as a marker for starting and ending points.  On these sequences, vision is mostly degraded by light absorption, strong illumination variations and backscattering.  In two sequences, visual information even becomes unavailable for a few seconds because of collisions with surrounding objects.  Another challenge is the presence of areas with seagrass moving because of the swell.  Moreover, the ROV is sensitive to waves and tether disturbances, which results in roll and pitch variations.

\subsection{Archaeological sites sequences}
\label{sec:archeo}

\begin{figure}
\centering
\includegraphics[width=0.95\linewidth]{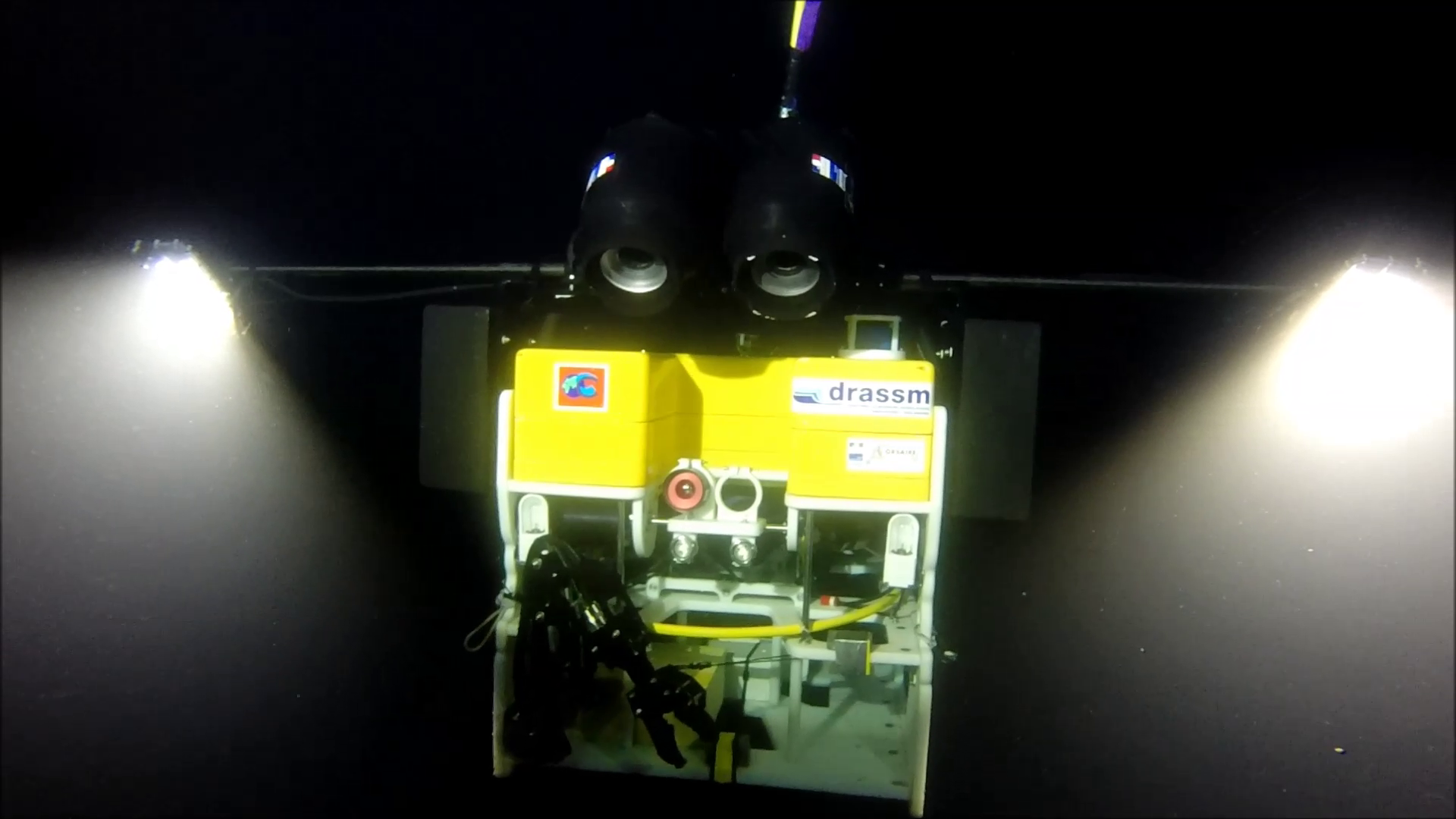}
\caption{The Remotely Operated Vehicle \textit{Perseo}, used on the archaeological sites. \\ \textit{Credit: F. Osada - DRASSM / Images Explorations}.}
\label{fig:perseo}
\end{figure}

The archaeological sites sequences were recorded in the Mediterranean sea, off Corsica's shore.  The System B, designed for deep waters, was embedded on the \textit{Perseo} ROV (\textit{Copetech SM Company}) displayed in Fig.~\ref{fig:perseo}.  In the way it was attached to the ROV, the camera viewing direction made a small angle with the vertical line ($\approx 20-30 \degree$).  \textit{Perseo} is equipped with two powerful led lights (250,000 lumens each) and with two robotics arms for manipulation purposes.  As localization while manipulating objects is a valuable information, to grab an artifact for instance, in some sequences the robotic arms are in the camera's field of view.  A total of 10 sequences have been recorded on these sites, with 3 sequences taken on the first site and 7 on the second one.  

\begin{figure}
    \centering
    \begin{subfigure}[b]{0.95\linewidth}
        \centering \includegraphics[width=0.95\linewidth]{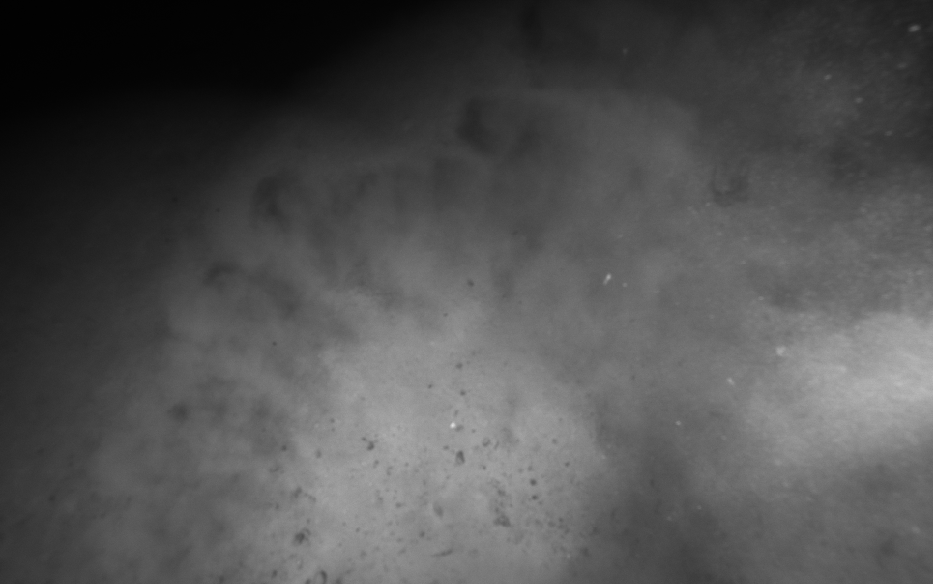}
        \captionsetup{justification=centering}
        \caption{Sandy cloud}\label{fig:archeo1_low}
    \end{subfigure}
    
    \begin{subfigure}[b]{0.95\linewidth}
        \centering \includegraphics[width=0.95\linewidth]{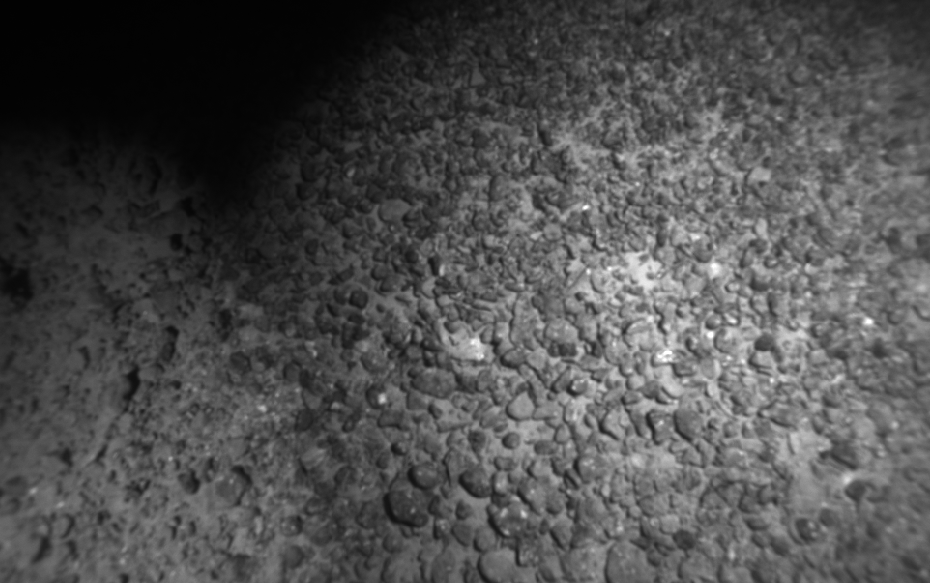}
        \captionsetup{justification=centering}
        \caption{Texture repetitive area}\label{fig:archeo1_rep}
    \end{subfigure}
    \captionsetup{justification=centering}
    \caption{Images acquired on the first archaeological site (depth: 270m).}
    \label{fig:archeo1}
\end{figure}

The first archaeological site explored was located at a depth of approximately 270 meters and hosted the remains of an antic shipwreck.  Hence, this site is mostly planar and presents mainly repetitive textures, due to numerous small rocks that were used as ballast in this antic ship (Fig.~\ref{fig:archeo1_low}).
These sequences are affected by turbidity and moving sand particles, increasing backscattering and creating sandy clouds (Fig.~\ref{fig:archeo1_rep}).  These floating particles are stirred up from the seabed by the water flows of the ROV's thrusters and lead to challenging visual conditions.  A shadow is also omnipresent in these sequences in the left corner of the recorded images, because of the limits of the lighting system.

\begin{figure}
    \centering
    \begin{subfigure}[b]{0.95\linewidth}
        \centering \includegraphics[width=0.95\linewidth]{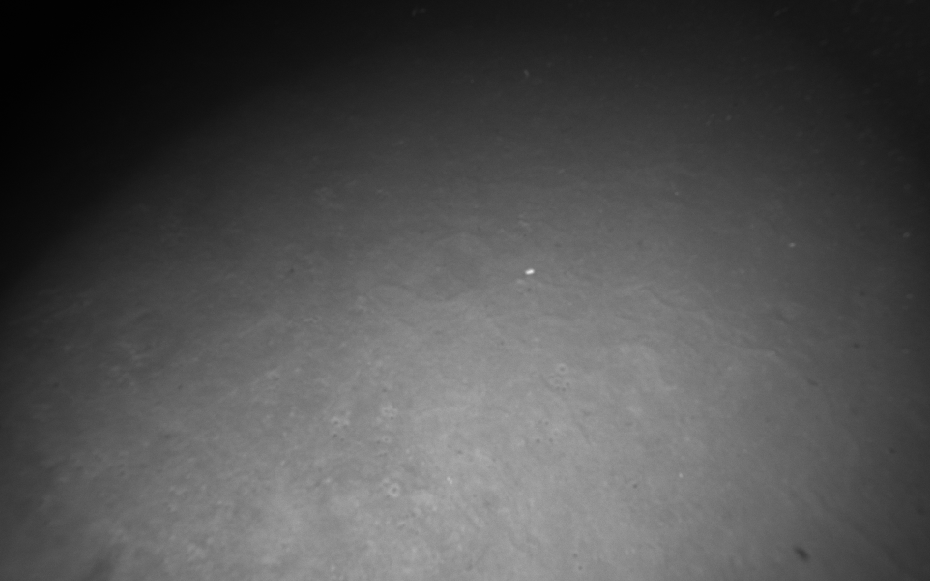}
        \captionsetup{justification=centering}
        \caption{Low texture area}\label{fig:archeo2_low}
    \end{subfigure}
    
    \begin{subfigure}[b]{0.95\linewidth}
        \centering \includegraphics[width=0.95\linewidth]{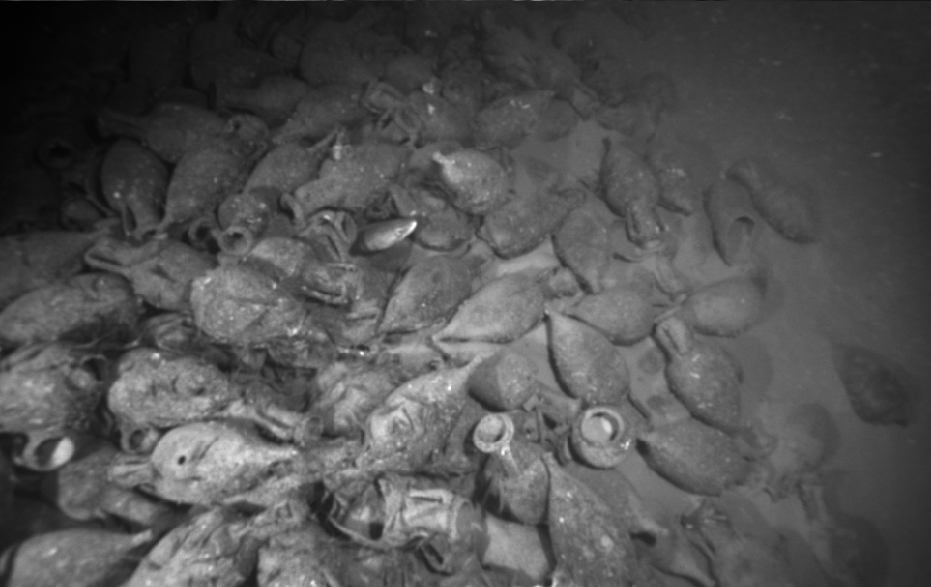}
        \captionsetup{justification=centering}
        \caption{Hill of amphorae}\label{fig:archeo2_amph}
    \end{subfigure}
    \captionsetup{justification=centering}
    \caption{Images acquired on the second archaeological site (depth: 380m).}
    \label{fig:archeo2}
\end{figure}

The second visited archaeological site was located at a depth of approximately 380 meters.  On this site a hill of amphorae is present (Fig.~\ref{fig:archeo2_amph}), whose top is culminating a few meters above the surrounding seabed level.  During these sequences, the ROV was mainly operated for manipulation and photogrammetry purposes.  While the amphorae present high texture, the ROV was also hovering low-textured sandy areas around the hill of amphorae (Fig.~\ref{fig:archeo2_low}).
Because of the presence of these amphorae, marine wildlife has been growing on this site.  Hence, the environment is quite dynamic, with many fishes getting in the field of view of the camera and many shrimps moving in the vicinity of the amphorae.  In one of the sequence, both arms get in front of the camera.  Otherwise, the visual degradation are the same as on the first site.

\begin{table*}
\centering
\begin{adjustbox}{width=1\textwidth}
\begin{tabular}{@{}cccccccccc@{}}
\toprule

\multirow{2}{*}{Site} & \multirow{2}{*}{Sequence}  &  \multirow{2}{*}{Duration}  &   \multirow{2}{*}{Length}   &   \multicolumn{6}{c}{Visual Disturbances}  \\ \cmidrule(l){5-10} 

& & & & 
Turbidity & Collisions & Backscattering & Sandy clouds & Dynamics & Robotic Arm \\

\midrule 
\midrule

\multirow{7}{*}{\shortstack{Harbor \\ \\ (depth $\approx$ 4 m) \\ \\ Acquired by system A, \\ embedded on a lightweight ROV}} 
                        &  \#01  & 3'49"  &   39.3m   &  X  &  -  &  X  &  -  &  -  &  - \\
                        &  \#02  & 6'47"  &   75.6m   &  X  &  -  &  X  &  -  &  -  &  - \\
                        &  \#03  & 4'17"  &   23.6m   &  X  &  -  &  X  &  -  &  -  &  - \\
                        &  \#04  & 3'26"  &   55.8m   &  X  &  X  &  X  &  -  &  -  &  - \\
                        &  \#05  & 2'52"  &   28.5m   &  X  &  -  &  X  &  -  &  -  &  - \\
                        &  \#06  & 2'06"  &   19.5m   &  X  &  -  &  X  &  -  &  -  &  - \\
                        &  \#07  & 1'53"  &   32.9m   &  X  &  X  &  X  &  -  &  -  &  - \\

\midrule 
\midrule

\multirow{5}{*}{\shortstack{First Archaeological Site \\ \\  (depth $\approx$ 270 m) \\ \\ Acquired by System B, \\ embedded on a medium workclass ROV}} 

                        &   &  &  &   &  &   &   &    &   \\
                        &  \#01  & 14'39"  &   32.4m   &  X  &  -  &  X  &  X  &  X  &  X \\
                        &  \#02  & 7'29"  &   64.3m   &  X  &  -  &  X  &  X  &  X  &  - \\
                        &  \#03  & 5'16"  &   10.7m   &  X  &  -  &  X  &  X  &  -  &  - \\
                        &   &  &  &   &  &   &   &    &   \\

\midrule 

\multirow{7}{*}{\shortstack{Second Archaeological Site \\ \\ (depth $\approx$ 380 m) \\ \\ Acquired by System B \\ embedded on a medium workclass ROV}}

                        &  \#04  & 11'09"  &   18.1m   &  X  &  -  &  X  &  X  &  X  &  X \\
                        &  \#05  & 3'19"  &   42.0m   &  X  &  -  &  X  &  -  &  X  &  - \\
                        &  \#06  & 2'49"  &   31.8m   &  X  &  -  &  X  &  -  &  X  &  - \\
                        &  \#07  & 9'29"  &   122.1m   &  X  &  -  &  X  &  -  &  X  &  - \\
                        &  \#08  & 7'49"  &   41.2m   &  X  &  -  &  X  &  -  &  X  &  - \\
                        &  \#09  & 5'49"  &   65.4m   &  X  &  -  &  X  &  -  &  X  &  - \\
                        &  \#10  & 11'54"  &   83.5m   &  X  &  -  &  X  &  -  &  X  &  - \\

\hline

\end{tabular}
\end{adjustbox}
\captionsetup{justification=centering}
\caption{Details on all the AQUALOC sequences and their associated visual disturbances.}
\label{tab:sequences}
\end{table*}

%\section{Ground truth}
\section{Comparative Baseline}
\label{sec:gt}

\begin{figure*}
    \centering
    \begin{subfigure}[b]{0.95\linewidth}
        \centering \includegraphics[width=0.975\linewidth]{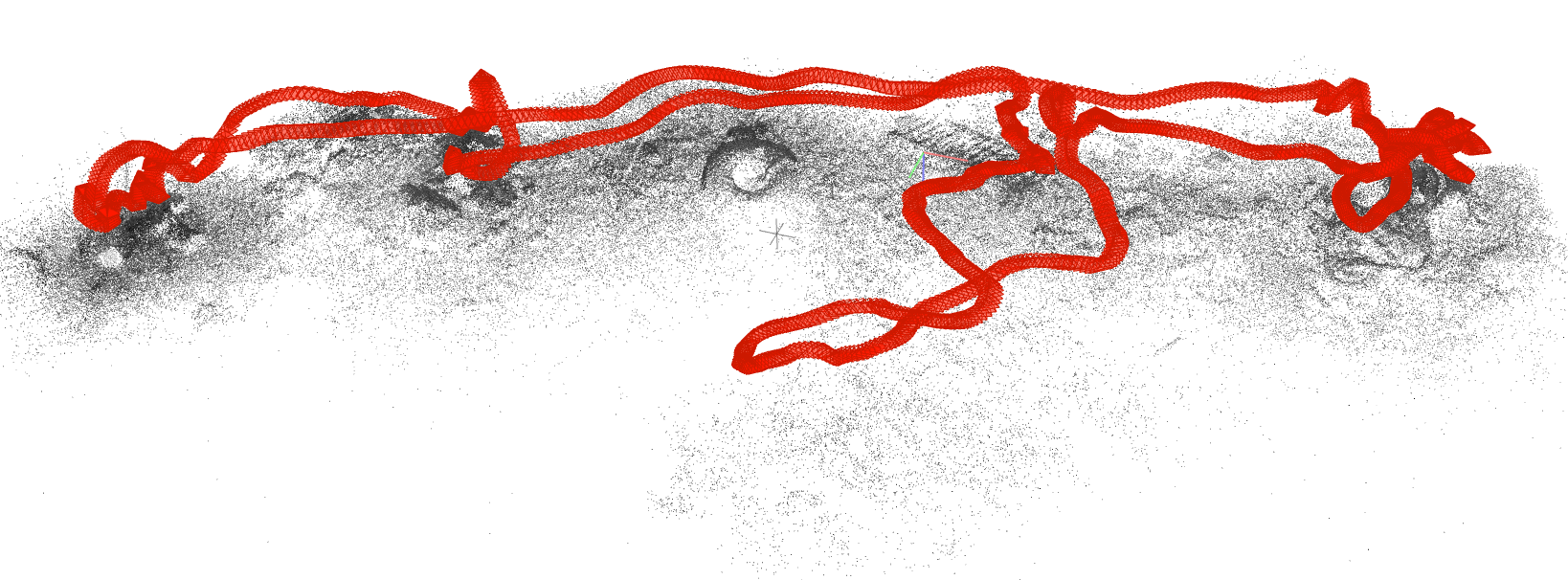}
        \captionsetup{justification=centering}
        \caption{Colmap reconstruction - Harbor \#02}\label{fig:colmap_1}
    \end{subfigure}
    
    \begin{subfigure}[b]{0.95\linewidth}
        \centering \includegraphics[width=0.975\linewidth]{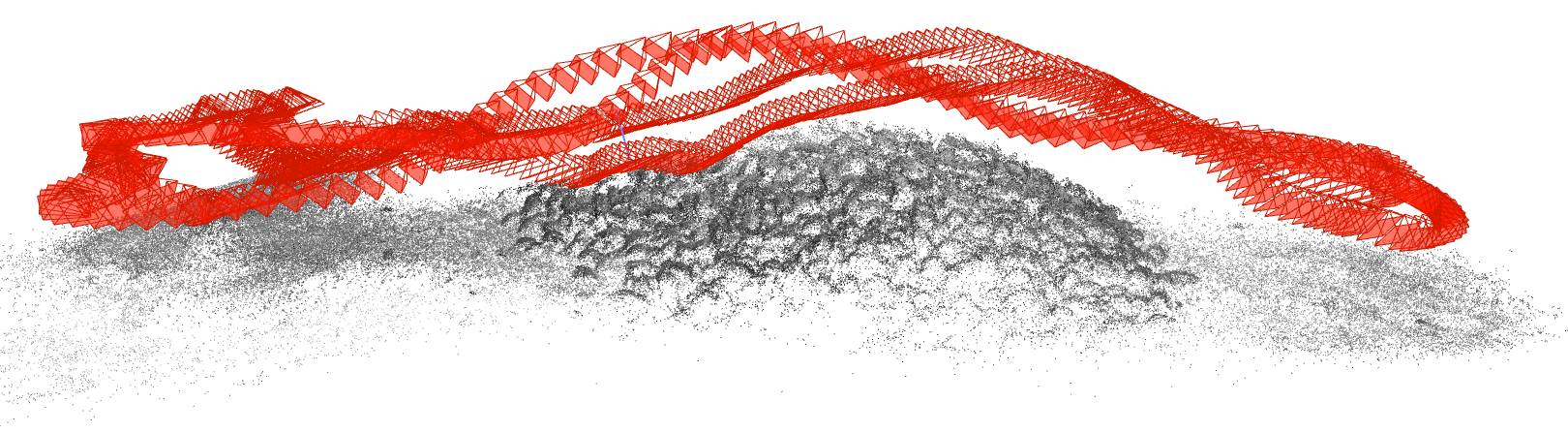}
        \captionsetup{justification=centering}
        \caption{Colmap reconstruction - Archeological Site \#07}\label{fig:colmap_1}
    \end{subfigure}
    
    \begin{subfigure}[b]{0.95\linewidth}
        \centering \includegraphics[width=0.9\linewidth]{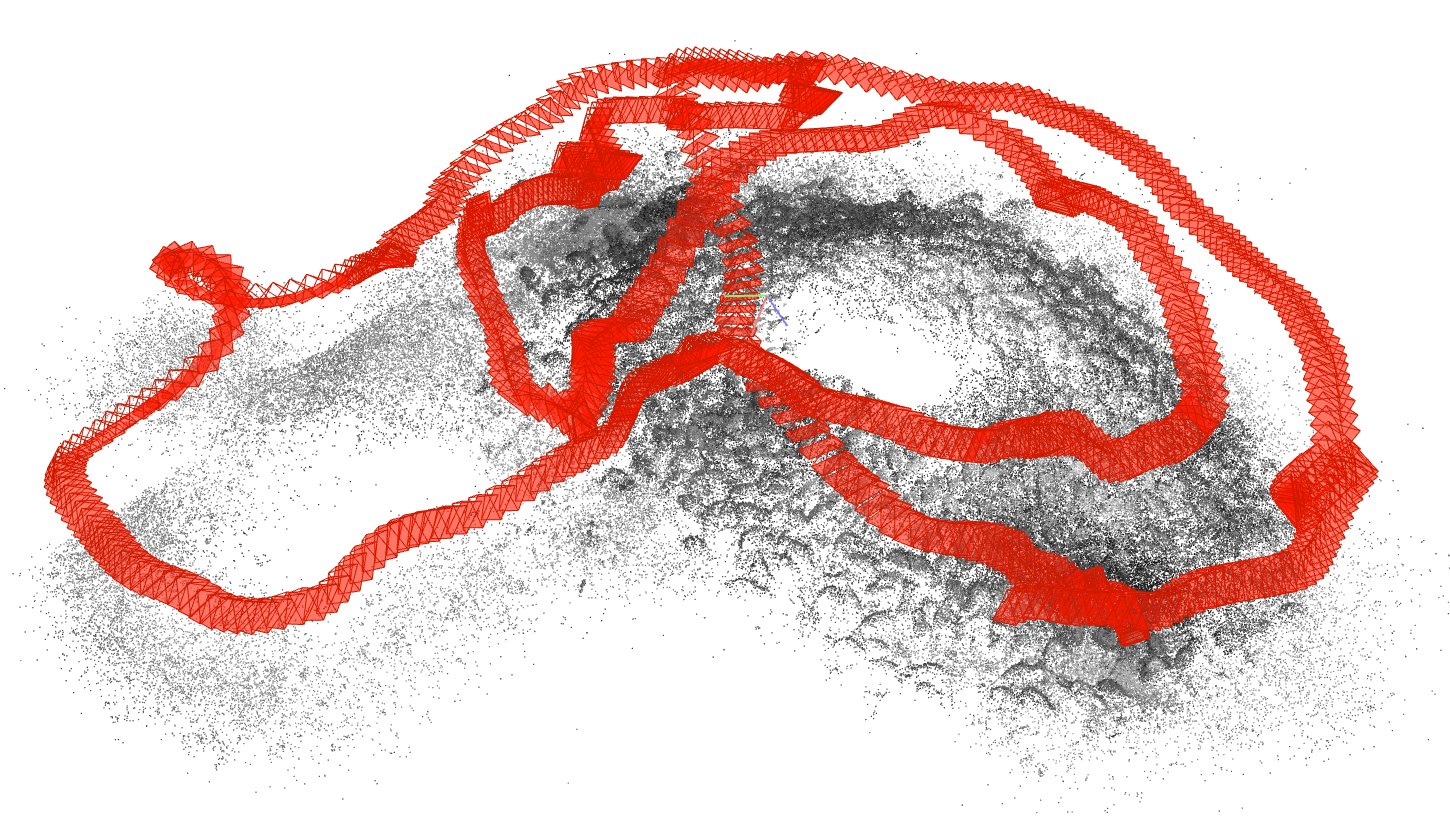}
        \captionsetup{justification=centering}
        \caption{Colmap reconstruction - Archeological Site \#10}\label{fig:colmap_1}
    \end{subfigure}
    
    \captionsetup{justification=centering}
    \caption{Examples of trajectories reconstructed with Colmap.}
    \label{fig:colmap}
    
    \vspace{2mm}
\end{figure*}

As the acquisition of a ground truth is very difficult in natural underwater environment, we have used the state-of-the-art Structure-from-Motion (SfM) software Colmap (\cite{Colmap}) to offline compute a 3D reconstruction for each sequence and extract a reliable trajectory from it.  By setting very low the features extraction parameters, we were able to extract enough SIFT features (\cite{Sift}) to robustly match the images of each sequence.  Performing a matching of the images in an exhaustive way, that is trying to match each image to all the other ones, allows to get a reliable trajectory reconstruction as many closed loops can be found (Fig.~\ref{fig:colmap}).  In Table~\ref{tab:colmap_stats}, we provide statistics for each sequence about Colmap's 3D reconstructions to highlight the reliability of the reconstructed models.  These statistics include the number of images used, the number of estimated 3D points, the average track length of each 3D points (\textit{i.e.} the number of images observing a given 3D point) and the average reprojection error.  The high average track lengths for each sequence (going from 6.7 to more than 20) assess the accuracy of the 3D points' estimation as it leads to a high redundancy in the bundle adjustment steps of the reconstruction.  Moreover, given these high track lengths, the average reprojection error is a good indicator of the overall quality of a SfM 3D model and for each one of the sequences this error is below 0.9 pixel.

The extracted trajectories have been scaled using the pressure sensor measurements and hence provide metric positions.  Although these trajectories cannot be considered as being perfect ground truths, we believe that it provides a fair baseline to evaluate and compare online localization methods.  Evaluation of such methods can be done using the standard Relative Pose Error (RPE) and Absolute Trajectory Error (ATE) metrics (\cite{Sturm_ATE}).  

Furthermore, we have made available the list of overlapping images (\textit{i.e.} matching) according to Colmap for each sequence.  These files could hence be used to evaluate the efficiency of loop-closure or image retrieval methods. 

%\begin{figure}
%\centering
%\includegraphics[width=0.95\linewidth]{colmap_seq10.png}
%\caption{Colmap reconstruction of the sequence \#10 of the archaeological sites.}
%\label{fig:colmap}
%\end{figure}

\section{Data Sequences Format}

As explained in the introduction, the sequences are all available as ROS bags and as raw data.  The dataset is split into two folders, one for the harbor sequences and the other for the archaeological ones.

\noindent The dataset repository architecture is the following:

\begin{forest}
  for tree={
    font=\ttfamily,
    grow'=0,
    child anchor=west,
    parent anchor=south,
    anchor=west,
    calign=first,
    edge path={
      \noexpand\path [draw, \forestoption{edge}]
      (!u.south west) +(7.5pt,0) |- node[fill,inner sep=1.25pt] {} (.child anchor)\forestoption{edge label};
    },
    before typesetting nodes={
      if n=1
        {insert before={[,phantom]}}
        {}
    },
    fit=band,
    before computing xy={l=15pt},
  }
[Harbor\_site\_sequences/
  [Calibration files/
    [camera\_calib.txt]
    [imu\_camera\_calib.txt]
    [imu\_noises.txt]
  ]
  [ground truth files/
    [colmap\_traj\_sequence\_X.txt]
    [...]
    [colmap\_detected\_loop\_sequence\_X.txt]
    [...]
  ]
  [sequence\_X\_bag.tar.gz/
    [sequence\_X.bag]
    [...]
  ]
  [sequence\_X\_raw\_data.tar.gz/
    [imu.csv]
    [mag.csv]
    [images.csv]
    [images/
      [frameXXXXX.png]
      [...]
    ]
  ]
]
\end{forest}

\noindent The archaeological sites sequences do not appear here but are organized exactly in the same manner.

The calibration files are given in the output format of Kalibr (\cite{Kalibr_cam,Kalibr_imucam}).

The trajectories computed by Colmap for each sequence are available as text files and contain the pose in a translation-quarternion form.  These files format is the following:

\begin{table}[!h]
\centering
\begin{adjustbox}{width=0.95\linewidth}
\begin{tabular}{@{}cccccccc@{}}
\textit{\#Frame} & \textit{tx} & \textit{ty} & \textit{tz} & \textit{qx} & \textit{qy} & \textit{qz} & \textit{qw} \\
 0. & -1.88 & 2.41 & -0.47 & 0.01 & 0.06 & 0.14 & 0.91 \\
20. & -1.83 & 2.35 & -0.46 & 0.05 & 0.64 & 0.14 & 0.99 \\
40. & -1.80 & 2.10 & -0.34 & 0.04 & 0.58 & 0.12 & 0.98 \\
...
\end{tabular}
\end{adjustbox}
\end{table}

The files containing the loop closures detected by Colmap provide information in the following format:
\begin{verbatim}
1,1,0,0,1
1,1,1,0,0
0,1,1,0,0
0,0,0,1,1
1,0,0,1,1
...
\end{verbatim}

\noindent where a $1$ indicates an overlapping between row $i$ and column $j$, with $i$ and $j$ standing for the frame numbers.  Note that only a subset of the images has been used to compute the offline reconstruction with Colmap (1 image out of 5 for the harbor sequences and 1 out 20 for the archaeological ones).  Therefore, the frame number given in these ground truth files is the number of their corresponding frame in the full sequence.

\begin{table*}
\centering
\begin{adjustbox}{width=1\textwidth}
\begin{tabular}{@{}lccccccccccccccccc@{}}
\toprule
                             & \multicolumn{7}{c|}{Harbor sequences}                                              & \multicolumn{10}{c}{Archeological sites sequences}                                    \\ \cmidrule(l){2-18} 
                             & \#01    & \#02    & \#03    & \#04    & \#05    & \#06    & \multicolumn{1}{c|}{\#07}    & \#01    & \#02    & \#03    & \#04    & \#05   & \#06   & \#07    & \#08    & \#09    & \#10    \\ \midrule
Nb. of used images             & 918    & 1590   & 1031   & 770    & 692    & 508    & \multicolumn{1}{c|}{447}    & 880    & 445    & 311    & 637    & 200   & 170   & 569    & 470    & 350    & 715    \\
Nb. of 3D points          & 112659 & 305783 & 355130 & 194407 & 236845 & 188807 & \multicolumn{1}{c|}{181964} & 196857 & 174514 & 160531 & 249048 & 42877 & 45799 & 251620 & 237882 & 114814 & 329686 \\
Mean tracking length          & 14.9   & 13.2   & 17.2   & 9.7    & 10.7   & 12.1   & \multicolumn{1}{c|}{9.5}    & 23.5   & 12.6   & 8.4    & 8.5    & 7.6   & 6.7   & 7.4    & 9.1    & 7.9    & 9.2    \\
Mean reproj. err. (px) & 0.896  & 0.816  & 0.713  & 0.715  & 0.688  & 0.733  & \multicolumn{1}{c|}{0.846}  & 0.746  & 0.621  & 0.474  & 0.673  & 0.601 & 0.569 & 0.645  & 0.616  & 0.660  & 0.661  \\ \bottomrule
\end{tabular}
\end{adjustbox}
\captionsetup{justification=centering}
\caption{Colmap trajectories reconstruction statistics.  The number of provided images, the number of reconstructed 3D points, the mean tracking length for the 3D points and the mean reprojection error for the 3D reconstruction are given for each sequence.}
\label{tab:colmap_stats}
\end{table*}

About the bag files, each sequence is stored in a separate bag containing the following topics:

\begin{itemize}
    \item \textbf{/camera/image\_raw}: Images recorded from the camera.
    \item \textbf{/camera/camera\_info}: Images width and height info.
    \item \textbf{/rtimulib\_node/imu}: Accelerometer and gyroscope measurements.
    \item \textbf{/rtimulib\_node/mag}: Magnetometer measurements.
    \item \textbf{/barometer\_node/pressure}: Pressure measurements in millibars.
    \item \textbf{/barometer\_node/depth}: Depth measurements in meters.
    \item \textbf{/barometer\_node/temperature}: Pressure sensor's temperature measurements.
\end{itemize}

In their raw format, each sequence contains the following data:

\begin{itemize}
    \item \textbf{}: The directory containing the sequence images.
    \item \textbf{frameXXXXX.png}: The images recorded from the camera.
    \item \textbf{images.csv}: The timestamps related to each image of the sequence.
    \item \textbf{imu.csv}: The accelerometer and gyroscope measurements and their timestamps.
    \item \textbf{mag.csv}: The magnetometer measurements and their timestamps.
    \item \textbf{depth.csv}: The pressure measurements converted in meters and their timestamps.
\end{itemize}

For each \textit{csv} files, the first row starts with a \# and then gives the name of the different fields along with their related measurements unit into squared brackets.  The following rows contain the values of the measurements.  In all these files, the first field is the acquisition timestamp of the measurements.  For instance, the depth.csv files look like:

\begin{verbatim}
#timestamp [ns], depth [m]
1542828791719540119,271.988866935
1542828791735507011,272.01910918
...
\end{verbatim}

\section{Conclusion}

In this paper, we have presented a new dataset of subsea monocular video sequences synchronized with inertial and pressure measurements.  This dataset is intended for encouraging the development of localization methods for underwater robots navigating close to the seabed.  The sequences have been recorded from Remotely Operated Vehicles in three different environments at different depths: a harbor at a depth of 4 meters, a first archaeological site at a depth of 270 meters and a second one at a depth of 380 meters.  The diversity of the recorded environments allowed to capture video sequences with different visual perturbations typical in underwater scenarios.  For each sequence, trajectories have been computed offline using a Structure-from-Motion library and are provided as a baseline for performance comparisons of localization methods.  The datasets are available both as ROS bags and as raw data.  In future work, we plan to perform new acquisition missions in different underwater environments in order to augment this dataset and increase its diversity.

\begin{acks}
The experiments conducted on the archaeological sites have been done in the French waters, in the framework of a research campaign of the DRASSM (French Department of Underwater Archaeology - Ministy of Culture) in accordance with the international regulation (UNESCO Convention on the Protection of the Underwater Cultural Heritage, 2001).

\noindent The authors are grateful to the DRASSM for its logistical support.

\noindent The authors acknowledge support of the CNRS (Mission pour l'interdisciplinarit\'e - Instrumentation aux limites 2018 - Aqualoc project) and support of R\'egion Occitanie (ARPE Pilotplus project).

\noindent The authors would also like to thank Anthelme Bernard-Brunel for his help in the design of the acquisition systems and Abderrahmane Kheddar for his helpful remarks.
\end{acks}

\subsection{References}

\bibliography{biblio.bib}
\bibliographystyle{SageH}

\end{document}